\documentclass{article}

\PassOptionsToPackage{numbers, compress}{natbib}
\usepackage[preprint]{neurips_2026}

\usepackage{array}
\usepackage[utf8]{inputenc}
\usepackage[T1]{fontenc}
\usepackage{hyperref}
\hypersetup{
   pdffitwindow=true,
   pdfstartview={FitH},
   pdfnewwindow=true,
   colorlinks,
   linktocpage=true,
   linkcolor=blue,
   urlcolor=blue,
   citecolor=blue
 }

\usepackage{caption}
\usepackage{url}
\usepackage{booktabs}
\usepackage{amsfonts}
\usepackage{nicefrac}
\usepackage{microtype}
\usepackage{xcolor}
\usepackage{graphicx}
\usepackage{amssymb}
\usepackage{amsmath}
\usepackage{makecell}
\usepackage{enumitem}
\usepackage{algorithm}
\usepackage{algpseudocode}
\usepackage{tabularx}
\usepackage{subcaption}
\usepackage{caption}
\usepackage{graphicx}       
\usepackage{booktabs}       
\usepackage{multirow}       
\usepackage{threeparttable} 
\usepackage{siunitx}        
\usepackage[capitalize,noabbrev]{cleveref}

\newcommand{\Z}{\mathcal{Z}}
\newcommand{\A}{\mathcal{A}}
\newcommand{\V}{\mathcal{V}}
\newcommand{\E}{\mathbb{E}}

\DeclareMathOperator{\ESS}{ESS}
\DeclareMathOperator{\argmax}{\arg \max}

\title{Self-Consistency via Marginal Sharpening}

\author{%
  Aleksei Arzhantsev\thanks{Equal contribution.} \\
  Criteo AI Lab, CREST IP Paris \\
  \texttt{a.arzhantsev@criteo.com} \\
  \And
  Otmane Sakhi\footnotemark[1] \\
  Criteo AI Lab \\
  \texttt{o.sakhi@criteo.com}
  \And
  Nicolas Chopin \\
  ENSAE, Institut Polytechnique de Paris \\
  \texttt{nicolas.chopin@cnrs.math.fr}
}

\begin{document}
\maketitle

\begin{abstract}
Inference-time sampling can elicit strong reasoning abilities from language models without additional training. Existing power-sampling methods do so by sharpening the distribution over full generated outputs, favoring completions that are individually likely under the model. We argue that this is the wrong object to target for reasoning: a completion entangles a reasoning trace with a final answer, whereas what matters is whether an answer is supported by many plausible reasoning paths. We therefore shift the target from the full-output distribution to the sharpened answer marginal, making self-consistency an inference-time objective rather than a post-hoc voting criterion. Surprisingly, this marginal target admits an efficient approximation: we propose a simple, purely autoregressive parallel sampling algorithm that approximately samples from the sharpened answer marginal, eliciting stronger performance than standard power sampling on mathematics and coding benchmarks while being orders of magnitude faster.
\end{abstract}

\section{Introduction}
\label{sec:intro}




Modern language models increasingly solve difficult tasks by allocating
generation to intermediate computation before producing a final response.
This behavior can be elicited at inference time with
chain-of-thought prompting \cite{wei2022chain, kojima2022large},
improved through rationale bootstrapping and process supervision
\cite{zelikman2022star, lightman2024lets}, and further shaped by
reinforcement learning on verifiable domains such as mathematics and code
\cite{guo2025deepseekr1, openai2026openaio1card}.  As a result, many current
models do not simply map a prompt directly to a short answer.  They generate a
latent trajectory, be it a derivation, program sketch, proof outline, or search
process, and only then emit the object that is evaluated.

This suggests a latent-variable view of reasoning models. For a given prompt,
there are many possible reasoning traces the model might follow, and each trace
induces a distribution over possible answers. A single sampled completion shows
only one such trace. It does not reveal the total support an answer receives
from the many alternative traces the model could have generated. From this
perspective, the relevant uncertainty is not just uncertainty over complete
outputs, but uncertainty over answers after aggregating across latent reasoning
paths.

A standard way to exploit this structure is to spend more computation at test
time. Search and sampling methods explore multiple continuations before
committing to a response, including beam-search variants for chain-of-thought
reasoning \cite{zhu2024deductive}, tree search \cite{yao2023tree}, Monte Carlo
tree search \cite{hao2023reasoning}, repeated sampling for code generation
\cite{chen2021evaluating}, and compute-scaling strategies for inference
\cite{snell2024scaling}. The simplest and most widely used method is
self-consistency: sample several reasoning traces and return the answer that
appears most often \cite{wang2022selfconsistency}. In the latent-variable view,
self-consistency works because it uses repeated samples to estimate which
answers are supported by many plausible reasoning paths.

However, majority voting is only a coarse way to use this idea. It works best
when answers are short, discrete, and easy to canonicalize. In more structured
domains, the same underlying solution can appear in many different forms. Two
programs may implement the same algorithm with different control flow, names,
or helper functions; two proofs may establish the same claim through different
lemmas; two derivations may reach the same result through different
intermediate steps. Exact voting over strings can fragment support across these
different realizations, even when they reflect the same answer-level belief.
This limitation has motivated heuristics that extend self-consistency beyond exact
string matching \cite{chen2024universal, cheng2025integrative}.

Recent power-sampling methods take a different approach. They sharpen the
distribution over complete generations, favoring outputs that are already
likely under the model as full sequences \cite{karan2025reasoning,
azizi2026powersmc, ji2026scalable}. For reasoning models, this means sharpening
a joint object: the reasoning trace together with the final answer. We argue
that this is the wrong level of abstraction for reasoning. The trace is a
latent computation, and many distinct traces may lead to the same answer. Sequence-level sharpening therefore treats alternative derivations,
implementations, or presentations as competitors, even when they support the
same answer-level belief. It concentrates probability on likely complete
transcripts rather than on answers with broad support across latent reasoning
paths.

We propose \emph{marginal sharpening}, a soft answer-level analogue of
self-consistency \cite{wang2022selfconsistency}. Rather than sharpening complete trajectories, marginal
sharpening sharpens the answer distribution obtained by aggregating over latent
reasoning traces. In simple discrete settings, this has the same intuition as
majority voting: answers reached by many plausible chains of thought should be
amplified. The difference is that marginal sharpening is an inference target,
not a post-hoc string-counting procedure. It does not require exact agreement
between sampled outputs, a small answer space, or an external canonicalizer.
This makes it suitable for code, proofs, and other structured generations,
where support for the same underlying solution may be spread across many
surface forms.

We study the resulting sharpened answer distribution and show that it admits a
useful decomposition, leading to a practical approximation with standard
autoregressive models. Our algorithm first samples multiple latent reasoning
traces, then decodes a single answer whose tokens are supported across those
traces. In this way, additional test-time compute is used to approximate the
sharpened answer marginal, rather than to search over complete trajectories or
vote over completed strings. We evaluate the approach on mathematical reasoning
and code-generation benchmarks, showing that marginal sharpening improves
final-answer performance from the same base models.

\paragraph{Contributions.}
\begin{itemize}
    \item We formulate \emph{marginal sharpening}, an answer-level inference-time
    objective for reasoning models that treats reasoning traces as latent
    variables and sharpens the induced answer distribution.

    \item We show that, for integer sharpening strengths, the sharpened answer
    marginal admits a useful multi-trace decomposition, connecting
    self-consistency to sampling from an answer-level target distribution.

    \item We propose a lightweight parallel autoregressive decoding algorithm
    that approximately samples from this sharpened distribution, and discuss an
    optional sequential importance sampling variant.

    \item Marginal sharpening substantially improves over
    power sampling on mathematics and coding benchmarks while running up to $38\times$ faster on
    long-reasoning problems. It achieves the best results on code generation
    benchmarks and remains close to majority voting on mathematical reasoning,
    where exact answer aggregation is a particularly strong baseline.
\end{itemize}

\paragraph{Outline.}
Section~\ref{sec:framework} presents the framework used throughout the paper. We describe the
autoregressive model, decompose each completion into a latent reasoning trace
and a final answer segment, define the induced answer marginal, and contrast
this object with sequence-level sharpening. Section~\ref{sec:marginal-sharpening} introduces marginal
sharpening. We define the answer-level target, derive the integer-strength
multi-trace representation, and turn this representation into practical
autoregressive decoding algorithms, including the lightweight parallel decoder
used in our main experiments. Section~\ref{sec:experiments}
evaluates the method on mathematical reasoning and code-generation benchmarks,
including comparisons to temperature sampling, majority voting, and power sampling,
as well as ablations over runtime, sharpening strength, approximations and fixed compute allocation.
Section~\ref{sec:conclusion} concludes. Appendix~\ref{app:results} reports the
full results table, and Appendix~\ref{app:sis} describes and evaluates the
optional sequential importance sampling correction.

\section{Framework}
\label{sec:framework}

\paragraph{Autoregressive language models.}
For a prompt $x \in \mathcal{X}$, an autoregressive language model defines a
distribution over complete token sequences $y=(y_1,\ldots,y_T) \in \mathcal{Y}$
by factorizing the probability of each sequence token by token:
\begin{equation}
    \pi(y \mid x)
    =
    \prod_{t=1}^{T} \pi(y_t \mid x, y_{<t}).
    \label{eq:ar}
\end{equation}
We include \texttt{EOS}, the end-of-sequence token in $y$, so the length $T$ is itself part
of the sampled sequence and may vary across samples.

\begin{figure}
    \centering
    \includegraphics[width=0.95\textwidth]{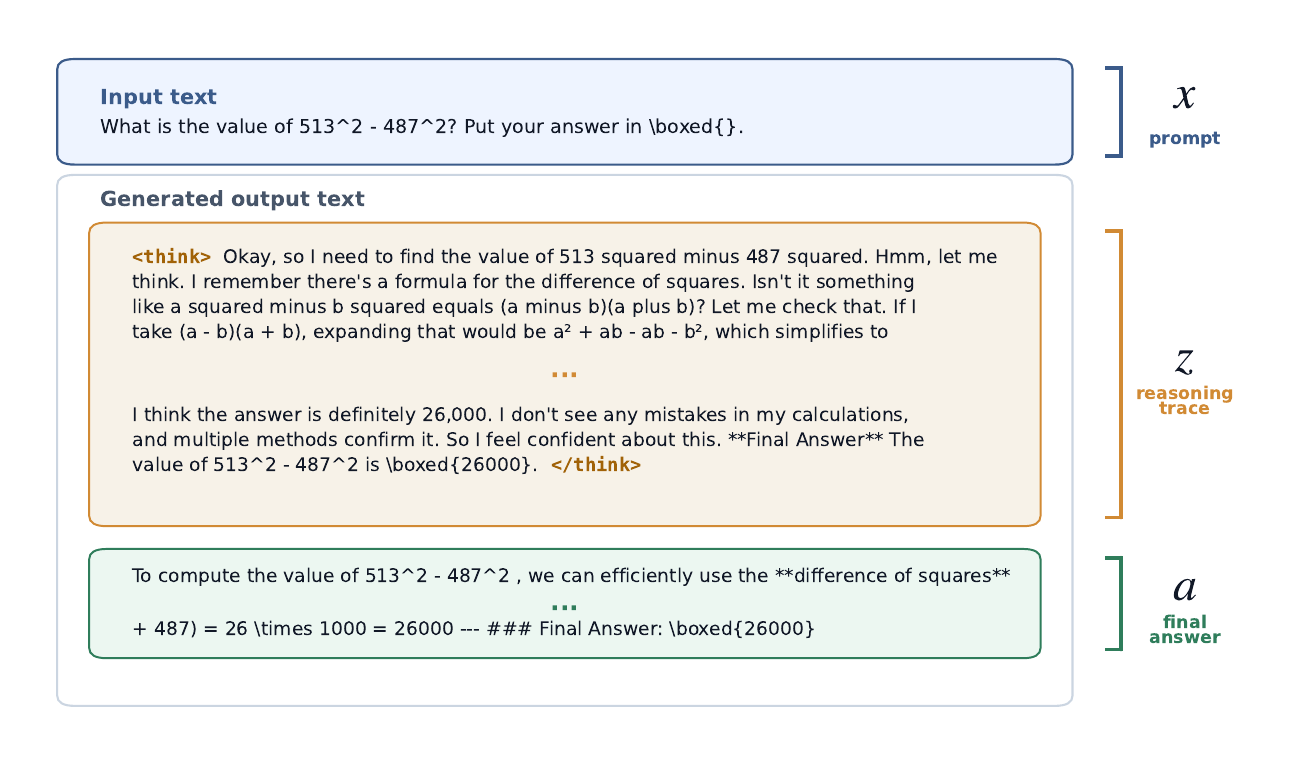}
    \caption{For a prompt \(x\), the reasoning-model completion decomposes into the reasoning trace \(z\), and final answer segment \(a\).  Everything inside the
    \texttt{<think>} and \texttt{</think>} tokens is considered a thinking trace, and everything that comes after the \texttt{</think>} token is considered as the final answer.}
    \label{fig:trace-answer-schema}
\end{figure}

\paragraph{Reasoning traces and answer segments.}
Reasoning-model completions naturally separate into two parts. A prefix
contains intermediate computation, such as scratch work, a derivation, a proof
outline, or a program sketch, while the suffix contains the final response
produced from that computation. We write the parsed completion as \(y=(z,a)\),
where \(z\) denotes the reasoning trace and \(a\) denotes the final answer
segment. This decomposition is only a parsing convention: the model still
assigns probability to the full sequence, and \(\pi(z,a\mid x)\) denotes the
probability of the sequence whose parsed components are \((z,a)\). 

In modern reasoning models \cite{yang2025qwen3, bakouch2025smollm3}, this split can often be
identified by a delimiter such as \texttt{</think>}\footnote{More generally,
the framework only requires a way to identify the answer segment.}.
Figure~\ref{fig:trace-answer-schema} illustrates the decomposition with an
example. The answer segment \(a\) may itself be a long structured sequence,
such as a program, a proof, or a natural-language final response.


With this distinction, the model induces a natural answer marginal by summing over all reasoning traces that lead
to the same answer string:
\begin{equation}
    m(a\mid x)
    =
    \sum_{z\in\Z}\pi(z,a\mid x)
    =
    \sum_{z\in\Z}\pi(z\mid x)\pi(a\mid x,z).
    \label{eq:answer-marginal}
\end{equation}



\paragraph{Power sampling.}
Power sampling \cite{karan2025reasoning} is a recent inference-time sharpening principle: instead of
sampling directly from the model distribution, it samples from a powered
version that amplifies high-probability generations. This gives the joint sharpening target
\begin{equation}
    p^{\mathrm{joint}}_\alpha(z,a\mid x)
    =
    \frac{\pi(z,a\mid x)^\alpha}{\sum_{z',a'}\pi(z',a'\mid x)^\alpha}.
    \label{eq:joint-sharpening}
\end{equation}
For $\alpha>1$, this distribution concentrates mass on complete trajectories
that are already likely under the base model. However, a complete trajectory contains
both a reasoning trace and the answer produced from it. Sequence-level power
sampling therefore sharpens the joint object \((z,a)\): it favors reasoning and answers that are likely together. Reasoning traces are best viewed as a latent computation that should be
aggregated over \cite{wang2022selfconsistency}. The same answer may be supported by many different
derivations, implementations, or proof strategies. Joint sharpening treats
these alternatives as competing complete sequences, rather than as shared
support for a common answer-level belief.


\section{Marginal Sharpening}
\label{sec:marginal-sharpening}



Treating reasoning as a latent process, we introduce \emph{Marginal sharpeningt}, that defines a power distribution over the answer marginal. For
sharpening strength \(\alpha>1\), the target is
\begin{equation}
    p^{\mathrm{marg}}_\alpha(a\mid x)
    =
    \frac{m(a\mid x)^\alpha}{Z_\alpha(x)},
    \qquad
    Z_\alpha(x)=\sum_{a'\in\A}m(a'\mid x)^\alpha.
    \label{eq:marginal-sharpening}
\end{equation}
The effect is to amplify answers with large total support across latent
reasoning traces. An answer can therefore receive high probability because it
is supported collectively by many plausible traces, even if no single complete
trajectory dominates. This is the answer-level analogue of power sampling:
the powered object is the marginal answer distribution, not the joint
distribution over traces and answers.

In the large-sharpening limit, the distribution concentrates on the most
supported answer:
\begin{equation}
    a \sim p^{\mathrm{marg}}_\alpha(\cdot \mid x)
    \xrightarrow[\alpha \rightarrow \infty]{}
    \argmax_{a'} m(a'\mid x).
\end{equation}
Thus, marginal sharpening is a soft version of self-consistency \cite{wang2022selfconsistency}: finite
\(\alpha\) samples from a sharpened answer distribution, while the limit
recovers the answer with maximum marginal support.





\paragraph{Integer-power trace representation.}
When the sharpening strength is an integer, \(\alpha=K\in\mathbb{N}\), the powered
answer marginal has an exact representation using \(K\) copies of the reasoning
trace variable.  Let \(\pi_z(\cdot\mid x)\) denote the distribution over reasoning
traces induced by the base model: operationally, we sample from \(\pi(\cdot\mid x)\)
but stop when the answer-boundary delimiter \texttt{</think>} is emitted.  After a trace \(z\) is fixed,
\(\pi(a\mid x,z)\) is the model probability of generating answer segment \(a\) as
the continuation.  Expanding the \(K\)th power gives
\begin{align}
    m(a\mid x)^K
    &=
    \left(\sum_z \pi_z(z\mid x)\pi(a\mid x,z)\right)^K \\
    &=
    \sum_{z_{1:K}}
    \prod_{i=1}^{K}\pi_z(z_i\mid x)\pi(a\mid x,z_i).
\end{align}
We rewrite it as expectation over $K$ independently sampled $z_i$.
\begin{align}
    m(a\mid x)^K
    =
    \sum_{z_{1:K}}
    \prod_{i=1}^{K}\pi_z(z_i\mid x)\pi(a\mid x,z_i)
    =
    \E_{z_{1:K}\sim\pi_z^{\otimes K}(\cdot\mid x)}
    \left[\prod_{i=1}^{K}\pi(a'\mid x,z_i)\right].
\end{align}
After normalization the final expression for the sharpened answer distribution becomes
\begin{align}
    p^{\mathrm{marg}}_K(a\mid x)
    &= \frac{
    \E_{z_{1:K}\sim\pi_z^{\otimes K}(\cdot\mid x)}
    \left[\prod_{i=1}^{K}\pi(a\mid x,z_i)\right]}
    {\sum_{a'\in\A}
    \E_{z_{1:K}\sim\pi_z^{\otimes K}(\cdot\mid x)}
    \left[\prod_{i=1}^{K}\pi(a'\mid x,z_i)\right]}.
    \label{eq:integer-expansion}
\end{align}

Equation~\eqref{eq:integer-expansion} gives the target in terms of trace samples:
for each candidate answer, we average its joint support across \(K\) independently
sampled reasoning traces and then normalize across answers.  This identity is the
theoretical object our algorithms approximate.  Using this would require
computing probabilities of complete answer sequences under each
\(\pi(a\mid x,z_i)\).  We therefore approximate this sequence-level conditional by
sampling autoregressively from a natural token-level distribution.


\paragraph{Trace-sample and token-level approximation.}
Equation~\eqref{eq:integer-expansion} expresses the sharpened answer marginal as an
expectation over \(K\) independently sampled reasoning traces.  We approximate this
expectation by drawing \(S\) independent \(K\)-trace groups
\(z^{(s)}_{1:K}\sim\pi_z^{\otimes K}(\cdot\mid x)\).  For complete answer strings,
the corresponding empirical approximation is
\begin{equation}
    \widehat p^{(S)}_K(a\mid x)
    =
    \frac{
    \sum_{s=1}^{S}\prod_{i=1}^{K}\pi(a\mid x,z_i^{(s)})}
    {\sum_{a'\in\A}\sum_{s=1}^{S}
    \prod_{i=1}^{K}\pi(a'\mid x,z_i^{(s)})}.
    \label{eq:rb-sequence}
\end{equation}

Directly sampling from Equation~\eqref{eq:rb-sequence} is still impractical because
\(\pi(a\mid x,z_i^{(s)})\) is the probability of a complete answer sequence.  To
see what an exact autoregressive sampler for Equation~\eqref{eq:rb-sequence} would
require, fix the current answer prefix \(c=a_{<t}=a_{\leq t-1}\).  The exact
next-token conditional is obtained by summing the sequence-level mass of all
complete answers that extend \(c\) with token \(v\):
\begin{equation}
    \widehat p^{(S)}_K(a_t=v\mid x,c)
    =
    \frac{
    \sum_{s=1}^{S}\sum_{r:\,cvr\in\A}
    \prod_{i=1}^{K}\pi(cvr\mid x,z_i^{(s)})
    }{
    \sum_{u\in\V}\sum_{s=1}^{S}\sum_{r:\,cur\in\A}
    \prod_{i=1}^{K}\pi(cur\mid x,z_i^{(s)})
    }.
    \label{eq:exact-prefix-conditional}
\end{equation}
Writing a complete answer as \(c,v,r\), where \(r\) denotes the remaining
continuation after token \(v\), the auto-regressive factorization gives
\[
    \prod_{i=1}^{K}\pi(cvr\mid x,z_i^{(s)})
    =
    \underbrace{
    \prod_{i=1}^{K}\pi(c\mid x,z_i^{(s)})
    }_{\text{prefix likelihood}}
    \underbrace{
    \prod_{i=1}^{K}\pi(v\mid x,z_i^{(s)},c)
    }_{\text{current-token score}}
    \underbrace{
    \prod_{i=1}^{K}\pi(r\mid x,z_i^{(s)},cv)
    }_{\text{continuation score}}.
\]
Therefore the exact conditional can be written as
\begin{equation}
    \widehat p^{(S)}_K(a_t=v\mid x,c)
    =
    \frac{
    \sum_{s=1}^{S}
    W_s(c)
    \prod_{i=1}^{K}\pi(v\mid x,z_i^{(s)},c)
    C_s(cv)
    }{
    \sum_{u\in\V}\sum_{s=1}^{S}
    W_s(c)
    \prod_{i=1}^{K}\pi(u\mid x,z_i^{(s)},c)
    C_s(cu)
    },
    \label{eq:exact-token-conditional}
\end{equation}
where
\[
    W_s(c)
    =
    \prod_{i=1}^{K}\pi(c\mid x,z_i^{(s)})
    =
    \prod_{\tau<t}\prod_{i=1}^{K}
    \pi(a_\tau\mid x,z_i^{(s)},a_{<\tau})
\]
is the likelihood assigned to the already generated prefix by trace group \(s\), and
\[
    C_s(cv)
    =
    \sum_{r:\,cvr\in\A}
    \prod_{i=1}^{K}\pi(r\mid x,z_i^{(s)},cv)
\]
is the continuation normalizer after choosing token \(v\).

The continuation normalizers \(C_s(cv)\) require summing over complete future
answers, which would defeat the goal of a lightweight decoder.  We therefore use a
local approximation that ignores the dependence of \(C_s(cu)\) on the candidate
token \(u\) and trace group \(s\), treating it as a common prefix-dependent factor.
This common factor cancels in the local normalization.  In contrast, the prefix
likelihood \(W_s(c)\) is already determined by the tokens generated so far and can
be maintained online.  We retain this term as a weight over sampled trace groups. For numerical stability, define the log-prefix weight
\[
    \ell_s(a_{<t})
    =
    \sum_{\tau<t}\sum_{i=1}^{K}
    \log \pi(a_\tau\mid x,z_i^{(s)},a_{<\tau}),
\]
and the normalized prefix weight
\[
    w_s(a_{<t})
    =
    \frac{\exp(\ell_s(a_{<t}))}
    {\sum_{s'=1}^{S}\exp(\ell_{s'}(a_{<t}))}.
\]
At \(t=1\), the prefix is empty, so \(\ell_s(\emptyset)=0\) and
\(w_s(\emptyset)=1/S\).  The resulting prefix-weighted token-level decoder is
\begin{equation}
    q_t^{(S)}(v\mid x,a_{<t})
    =
    \frac{
    \sum_{s=1}^{S}
    w_s(a_{<t})
    \exp\left(
    \sum_{i=1}^{K}
    \log \pi(v\mid x,z_i^{(s)},a_{<t})
    \right)}
    {\sum_{u\in\V}\sum_{s=1}^{S}
    w_s(a_{<t})
    \exp\left(
    \sum_{i=1}^{K}
    \log \pi(u\mid x,z_i^{(s)},a_{<t})
    \right)}.
    \label{eq:poe-token-advanced}
\end{equation}
After sampling \(a_t\sim q_t^{(S)}(\cdot\mid x,a_{<t})\), the log-prefix weights
are updated by
\[
    \ell_s(a_{\leq t})
    =
    \ell_s(a_{<t})
    +
    \sum_{i=1}^{K}
    \log \pi(a_t\mid x,z_i^{(s)},a_{<t}).
\]
The decoder then appends \(a_t\) to the shared answer prefix and repeats until the
answer terminates or reaches the length budget.  This rule favors tokens that are
likely under several independently sampled reasoning traces, while giving greater
influence to trace groups that have remained consistent with the answer prefix
generated so far.  It avoids enumerating complete answer strings and replaces the
intractable continuation normalizers by a local approximation. With this token level proposal, we define Marginal Sharpening in Algorithm~\ref{alg:marginal-sharpening}.

The local rule in Equation~\eqref{eq:poe-token-advanced} can be of obviously improved.  A more accurate sampler could spend additional compute on full-sequence corrections, for example by applying Metropolis--Hastings updates as in power sampling \cite{karan2025reasoning}, or by using SMC-style importance corrections and resampling as in Power-SMC \cite{azizi2026powersmc}. These approaches can better account for the missing continuation normalizers, but require multiple proposed continuations or repeated resampling steps.  We keep the lightweight token-level decoder as the main algorithm and discuss heavier corrections in Appendix~\ref{app:sis}.

\paragraph{Cost and implementation notes.}
The algorithm has two knobs. The sharpening strength \(K\) controls how
strongly the answer distribution is concentrated, while the sample count \(S\)
controls the quality of the token-level approximation. Generating each answer
token requires \(\mathcal{O}(KS)\) model evaluations, but these evaluations are
parallel across trace groups, batched in practice, and reuse cached states.
When fewer than two usable reasoning traces are available, we fall back to the
corresponding base completion rather than applying the multi-trace decoder.


\begin{algorithm}
\caption{Marginal Sharpening}
\label{alg:marginal-sharpening}
\begin{algorithmic}[1]
\Require Prompt \(x\), sharpening strength \(K\), trace groups \(S\), maximum
output length \(L\)
\State Generate \(M=K\cdot S\) reasoning traces for \(x\), i.e. for all
\(m\in[M]\), \(z_m\sim\pi_z(\cdot\mid x)\).
\State Form \(S\) groups \(z^{(s)}_{1:K}\), each containing \(K\) traces.
\State Initialize the answer \(a_{\leq 0}\gets\emptyset\) and prefix log-weights
\(\ell_s\gets 0\) for all \(s\in[S]\).
\State Compute remaining budget $ T\gets L-\sum_{s,k}|z_k^{(s)}|/M .$
\For{\(t=1,\ldots,T\)}
    \State Compute the prefix-weighted next-token distribution
    \Statex \[
        q_t^{(S)}(v\mid x,a_{\leq t-1})
        =
        \frac{
        \sum_{s=1}^{S}\exp\left(
        \ell_s+
        \sum_{i=1}^{K}\log \pi(v\mid x,z_i^{(s)},a_{\leq t-1})\right)}
        {\sum_{u\in\V}\sum_{s=1}^{S}\exp\left(
        \ell_s+
        \sum_{i=1}^{K}\log \pi(u\mid x,z_i^{(s)},a_{\leq t-1})\right)}.
    \]
    \State Sample \(a_t\sim q_t^{(S)}(\cdot\mid x,a_{\leq t-1})\).
    \State Update prefix log-weights for all \(s\in[S]\):
    \[
        \ell_s \gets \ell_s+
        \sum_{i=1}^{K}\log \pi(a_t\mid x,z_i^{(s)},a_{\leq t-1}).
    \]
    \State Append the new token to the answer \(a_{\leq t}\gets(a_{\leq t-1},a_t)\).
    \State \textbf{If} {\(a_t=\texttt{EOS}\)} \textbf{then} \texttt{break} .
\EndFor
\State \Return reasoning traces \(\{z_m\}_{m\in[M]}\) and decoded answer \(a_{\leq t}\).
\end{algorithmic}
\end{algorithm}

\section{Related Work}
\label{sec:related}

\textbf{Chain-of-thought and reasoning models.}
Chain-of-thought prompting showed that language models can improve on
multi-step tasks by generating intermediate reasoning before the final answer
\cite{wei2022chain}, and similar behavior can be elicited with simple
zero-shot prompts \cite{kojima2022large}. Recent reasoning models make this
trace--answer structure central, often using post-training or reinforcement
learning to produce longer reasoning traces for mathematics and code
\cite{guo2025deepseekr1,kimi2025k15}. Our framework uses this structure
directly, while remaining agnostic to how the traces are produced.

\textbf{Self-consistency.}
Self-consistency uses repeated sampling to estimate answer-level support:
sample multiple reasoning paths, extract their final answers, and return the
answer with the largest vote count \cite{wang2022selfconsistency}. This works
well when answers can be exactly matched or canonicalized, but becomes brittle
when the same solution has many surface forms. Universal self-consistency and
integrative decoding address this limitation with model-based aggregation or
implicit self-consistency \cite{chen2024universal,cheng2025integrative}.
Marginal sharpening takes a different route: it makes the answer marginal
itself the inference target. Under this view, self-consistency is a limiting
case, while finite sharpening gives a soft answer-level sampler.

\textbf{Inference-time sharpening.}
Power-sampling methods sharpen model distributions at inference time without
additional training. Reasoning with Sampling uses Metropolis--Hastings updates
to sample from sequence-level power distributions
\cite{karan2025reasoning}; Scalable Power Sampling uses an autoregressive
proposal with future-continuation corrections \cite{ji2026scalable}; and
Power-SMC uses sequential importance corrections and resampling for the same
trajectory-level target \cite{azizi2026powersmc}. These methods sharpen
complete generations. Marginal sharpening instead changes the object of
sharpening: it sharpens the answer marginal induced by latent reasoning traces.

\section{Experiments}
\label{sec:experiments}

\textbf{Datasets.}
We evaluate on three answer-focused reasoning settings. MATH-500 is a subset of the
MATH benchmark for competition-style mathematical problem solving \cite{hendrycks2021math}; answers are
extracted from boxed final expressions and checked symbolically with the Math-Verify
library \cite{mathverify2025}. HumanEval evaluates Python function synthesis from
natural-language docstrings using unit tests \cite{chen2021codex}; we use
HumanEval+, which adds stronger tests for the same problems \cite{liu2023evalplus}.
LiveCodeBench v6 is a more challenging contest-style code generation benchmark built
from recent programming problems and evaluated with the official runner
\cite{jain2024livecodebench}.

\textbf{Models.}
We use the 1.7B, 4B and 8B version of the Qwen family \cite{yang2025qwen3} in our experiments.  These models emit an explicit reasoning segment before
the final answer.  In our implementation, the reasoning trace is the generated prefix
ending at \texttt{</think>}; the continuation after this delimiter is treated as the
answer segment.  Unless stated otherwise, all sampling uses temperature 1.0.

\textbf{Main experiment.} We present the main results of our experiment in Table~\ref{tab:main-results-l8192}. All results in this table use a maximum generation length of \(L=8192\). Power Sampling \cite{karan2025reasoning} is used with its default parameters ($\alpha = 4, B=16, N=10$). For Qwen3-8B, power sampling was too slow ($>4$ hours a prompt) and did not finish . For each model and dataset, the best result is shown in bold and the second-best result is underlined.  Marginal sampling gives the best results on both coding benchmarks, where many different programs can implement the same solution and direct voting over final strings is less effective.  On MATH-500, marginal sampling is second best, while majority voting performs best.  This is expected: mathematical answers have a much smaller effective answer space, and equivalent expressions can be compared symbolically, making majority voting a strong baseline.

\begin{table}
\centering
\small
\caption{Accuracy across 3 benchmarks. Best results are bolded; second best are underlined.}
\label{tab:main-results-l8192}
\begin{tabular}{llccc}
\toprule
Model & Method & MATH-500 & HumanEval+ & LiveCodeBench \\
\midrule
Qwen3-1.7B & Temperature Sampling & 0.806 & 0.722 & 0.227 \\
 & Power Sampling & 0.762 & 0.713 & 0.234 \\
 & Majority Voting ($k = 32$) & \textbf{0.920} & \underline{0.756} & 0.229 \\
 \cmidrule(lr){2-5}
 & Marginal Sharpening $(K=4, S=8)$ & 0.904 & \textbf{0.762} & \underline{0.269} \\
 & Marginal Sharpening $(K=32, S=1)$ & \underline{0.908} & \underline{0.756} & \textbf{0.280} \\
\midrule
\midrule
Qwen3-4B & Temperature Sampling & 0.709 & 0.843 & 0.266 \\
 & Power Sampling & 0.694 & 0.890 & \underline{0.303} \\
 & Majority Voting ($k = 32$) & \textbf{0.848} & 0.884 & 0.274 \\
 \cmidrule(lr){2-5}
 & Marginal Sharpening $(K=4, S=8)$ & \underline{0.842} & \underline{0.915} & \textbf{0.309} \\
 & Marginal Sharpening $(K=32, S=1)$ & 0.840 & \textbf{0.927} & \underline{0.303} \\
\midrule
\midrule
Qwen3-8B & Temperature Sampling & 0.845 & 0.840 & 0.296 \\
& Power Sampling & - & - & - \\
 & Majority Voting ($k = 32$) & \textbf{0.932} & 0.854 & 0.309 \\
 \cmidrule(lr){2-5}
 & Marginal Sharpening $(K=4, S=8)$ & \underline{0.928} & \underline{0.860} & \textbf{0.349} \\
 & Marginal Sharpening $(K=32, S=1)$ & 0.924 & \textbf{0.878} & \underline{0.343} \\
\bottomrule
\end{tabular}
\end{table}


\textbf{Comparison to power sampling.} Marginal sharpening is better suited to long generations than power sampling in the regime we consider.  Power sampling repeatedly resamples parts of the same output sequence to approximate the sequence-level power distribution.  With maximum generation length \(L\), this requires \(O(L^2)\) generated tokens per problem.  In contrast, for fixed \(K\) and \(S\), marginal sharpening decodes autoregressively and requires \(O(L)\) generated tokens per problem.  The reasoning traces used by marginal sharpening are also sampled independently, so trace generation can be parallelized directly.

\begin{figure}
    \centering
    \includegraphics[width=0.9\textwidth]{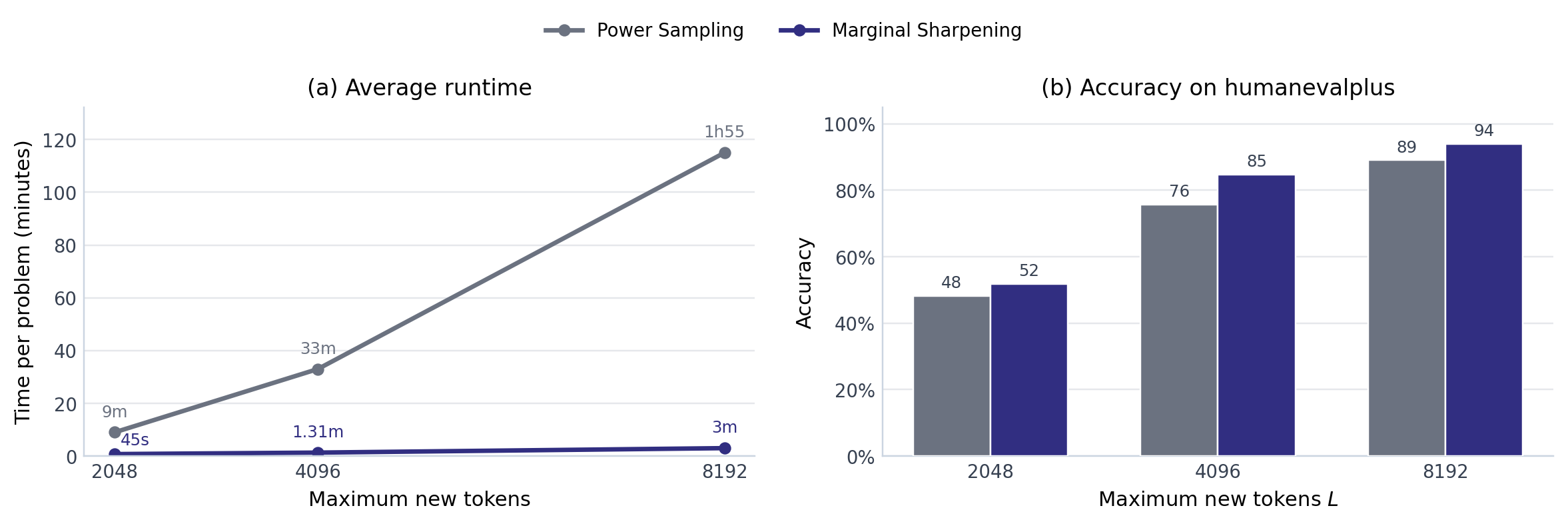}
    \caption{Runtime and accuracy comparison for Qwen3-4B on HumanEval+ as the maximum generation length changes.}
    \label{fig:runtime-accuracy}
    \vspace{-0.5cm}
\end{figure}

Figure~\ref{fig:runtime-accuracy} compares runtime and accuracy for Qwen3-4B on HumanEval+ as \(L\) varies. Running both methods on $1$ H100 GPU, Marginal sharpening takes about 45 seconds, 1.31 minutes, and 3 minutes per problem for \(L=2048,4096,8192\), respectively, compared with 9, 33, and 115 minutes for power sampling.  This is roughly a \(12\times\), \(25\times\), and \(38\times\) reduction in runtime.  Despite this difference, marginal sharpening obtains better accuracy, showing that answer-level sharpening captures much of the useful effect of power sampling at a substantially lower computational cost.

\textbf{Effect of the sharpening strength.} We next study the effect of the integer sharpening strength \(K\).  In our
integer-power representation, \(K\) is both the exponent of the sharpened answer
marginal and the number of sampled reasoning traces used by the decoder.
Figure~\ref{fig:qwen-K-impact-results} shows Qwen3-8B results across all three
datasets for several values of \(K\).  The leftmost bar, temperature sampling, draws directly from the base model and can
be viewed as the unsharpened \(K=1\) case.  Increasing \(K\)
generally improves performance, and \(K=32\) gives the best results in this sweep.
This is consistent with the role of \(K\): larger values put more weight on answers
that are supported by several traces, making the answer distribution sharper.  This
can improve accuracy when the model assigns substantial marginal mass to the correct
answer.  However, larger \(K\) also requires more computation and can reduce diversity,
so smaller values may be preferable when optimizing the quality--compute tradeoff.



\begin{table}
\centering

\begin{minipage}[t]{0.6\textwidth}
    \centering
    \vspace{0pt}
    \includegraphics[width=\linewidth]{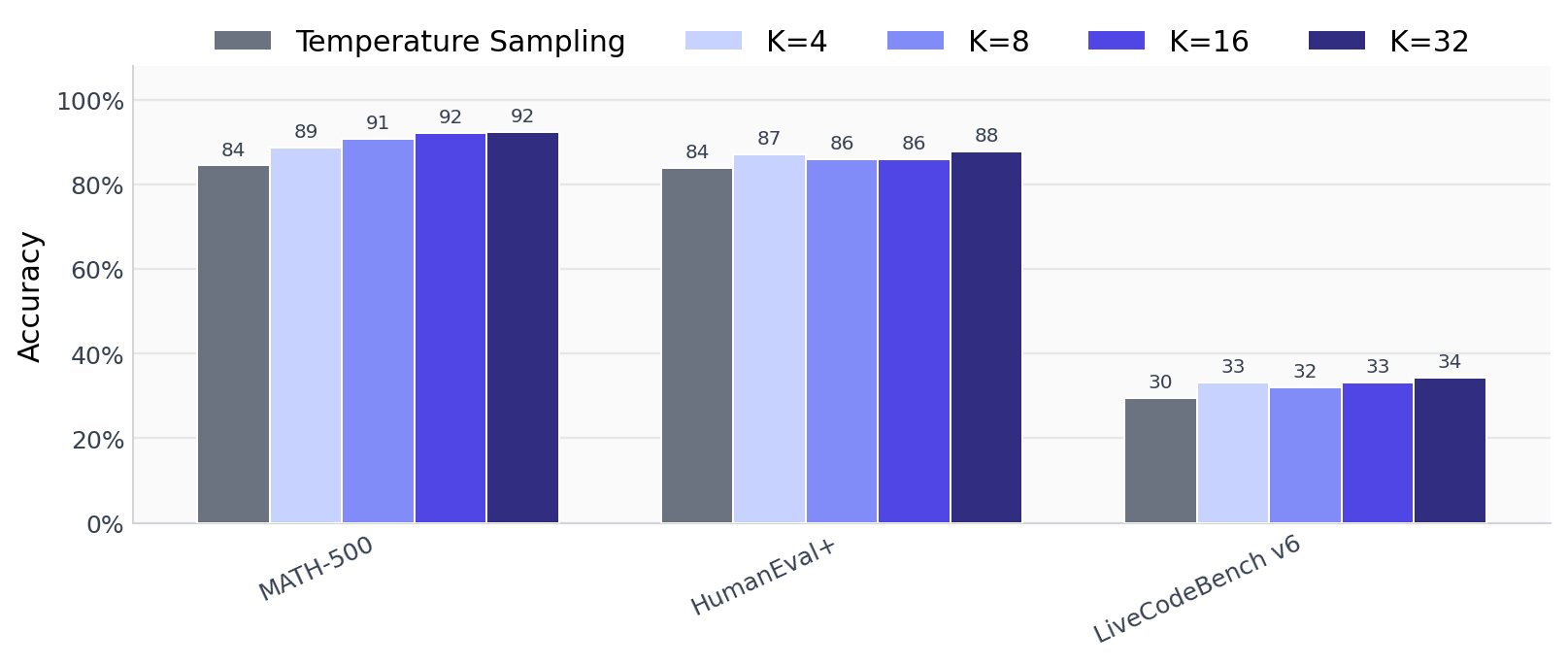}
    \captionof{figure}{Qwen3-8B results on MATH-500, HumanEval+, and
    LiveCodeBench for different sharpening strengths \(K\).}
    \label{fig:qwen-K-impact-results}
\end{minipage}
\hfill
\begin{minipage}[t]{0.39\textwidth}
    \centering
    \vspace{0pt}
    \scriptsize
    \caption{Qwen3-4B: Comparison of marginal sharpening varying $S$.}
    \label{tab:qwen3_s18}
    \setlength{\tabcolsep}{3pt}
    \begin{tabular}{ccccc}
    \toprule
    \(L\) & \(S\) & MATH-500 & HumanEval+ & LCB v6 \\
    \midrule
    2048 & 1 & 0.410 & 0.317 & 0.114 \\
    2048 & 8 & \textbf{0.522} & \textbf{0.512} & \textbf{0.131} \\
    \midrule
    4096 & 1 & 0.560 & 0.695 & 0.189 \\
    4096 & 8 & \textbf{0.702} & \textbf{0.829} & \textbf{0.200} \\
    \midrule
    8192 & 1 & 0.778 & \textbf{0.915} & 0.303 \\
    8192 & 8 & \textbf{0.842} & \textbf{0.915} & \textbf{0.309} \\
    \bottomrule
    \end{tabular}
\end{minipage}

\end{table}

\textbf{Effect of Monte Carlo Samples $S$.} Table~\ref{tab:qwen3_s18} compares two values of \(S\), while keeping the model, \(K\), and length budget fixed. Increasing \(S\) gives consistently stronger results for Qwen3-4B:
\(S=8\) improves over \(S=1\) on MATH-500 and LiveCodeBench v6 for every length
budget, and improves or matches it on HumanEval+.  The gains are especially visible
at smaller budgets, suggesting that a more accurate approximation can partly
compensate for limited generation length.

\begin{figure}
    \centering
    \includegraphics[width=0.95\textwidth]{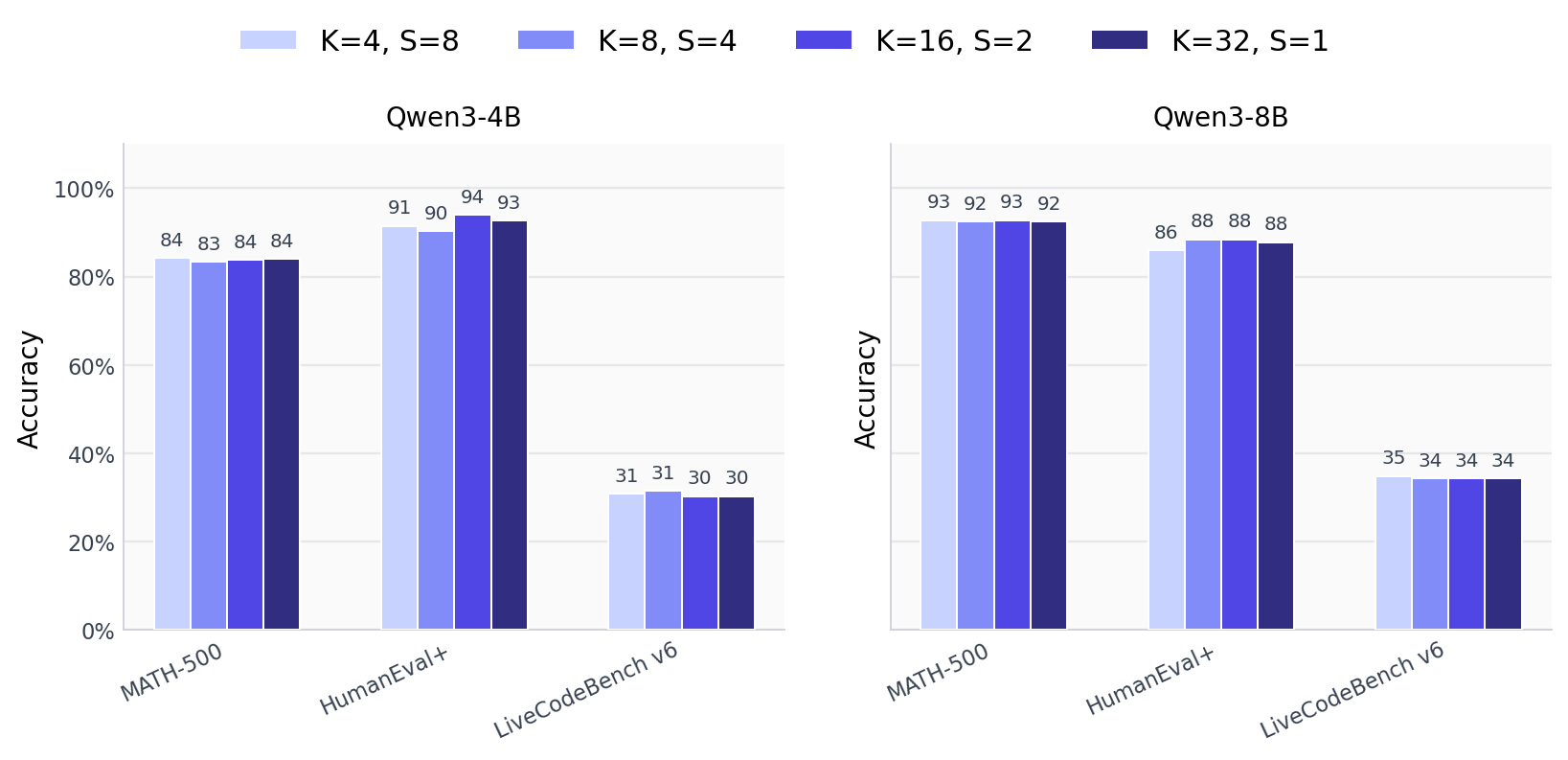}
    \caption{Comparison of marginal sharpening configurations with the same total
    generation budget \(K\cdot S=32\).  Rows correspond to length budgets and columns to
    Qwen3 model sizes.}
    \label{fig:budget32-ks-grid}
    \vspace{-0.5cm}
\end{figure}

\textbf{Fixed-budget allocation between \(K\) and \(S\).} We also compare different ways to allocate the same budget of 32 generated traces.
Figure~\ref{fig:budget32-ks-grid} shows all configurations with \(K \cdot S=32\).  This
comparison separates two uses of compute within the same sampling family.  Larger
\(K\) sharpens the answer distribution,
while larger \(S\) spends more computation on approximating the conditional
distribution for a fixed trace set.  The results show that no single allocation
dominates across all settings.  Instead, the configurations are mostly close to each
other across datasets, length budgets, and model sizes.  This suggests that, at
fixed total compute, the two effects largely cancel out: increasing \(K\) sharpens
the target more aggressively, while increasing \(S\) gives a better local
approximation to that target.

\textbf{Summary.} The experiments support marginal sharpening as a strong test-time inference
method and reinforce the central intuition behind self-consistency: for
reasoning models, it is often better to sharpen the answer distribution induced
by latent traces than to sharpen the distribution over complete trajectories.
This distinction is especially important for code generation, where many
syntactically different programs can implement the same solution and exact
majority voting becomes brittle. Marginal sharpening avoids requiring a
canonical final string while still using multiple reasoning traces to
concentrate probability on answers that are jointly supported by the model.

The ablations suggest that the main gains come from the answer-level target and
from using multiple traces, rather than from exactly correcting every
sequence-level conditional. Increasing \(K\) and \(S\) generally improves
performance, but under a fixed generation budget, different allocations between
sharpening strength and sample count are often close. This suggests that both
stronger sharpening and better approximation quality are useful, with no single
allocation dominating across all settings.

\section{Conclusion}
\label{sec:conclusion}


We introduced \emph{marginal sharpening}, an inference-time method that treats
reasoning traces as latent variables and sharpens the answer marginal they
induce, rather than the probability of complete trajectories. For integer
sharpening strengths, this target admits a multi-trace representation, yielding
a simple autoregressive decoder that samples several traces and produces one
answer supported across them. Empirically, marginal sharpening improves over
temperature sampling, outperforms sequence-level power sampling in the
long-generation regime while running up to \(38\times\) faster, and is
especially strong on code-generation benchmarks where exact string agreement is
too restrictive. These results show that self-consistency can be understood
as an answer-level inference principle than as a voting heuristic, and that
test-time compute is more effectively spent on the answer belief induced by
latent reasoning than on selecting the most likely transcript. Marginal sharpening is only one approximation to this principle:
developing better samplers for the sharpened answer marginal, with tighter
corrections for the token-level approximation are promising
directions for future work.


\section{Acknowledgment}
\label{sec:ack}
This project was provided with computational AI and storage
resources by GENCI at IDRIS thanks to the grant 2025-A0191016862 on the supercomputer Jean Zay
(V100/A100/H100 partitions).

\bibliographystyle{unsrt}
\bibliography{reference}

@article{karan2025reasoning,
  title={Reasoning with Sampling: Your Base Model is Smarter Than You Think},
  author={Karan, Aayush and Du, Yilun},
  journal={arXiv preprint arXiv:2510.14901},
  year={2025},
  url={https://arxiv.org/abs/2510.14901}
}

@article{wei2022chain,
  title={Chain-of-Thought Prompting Elicits Reasoning in Large Language Models},
  author={Wei, Jason and Wang, Xuezhi and Schuurmans, Dale and Bosma, Maarten and Ichter, Brian and Xia, Fei and Chi, Ed and Le, Quoc and Zhou, Denny},
  journal={Advances in Neural Information Processing Systems},
  volume={35},
  pages={24824--24837},
  year={2022}
}

@misc{openai2026openaio1card,
      title={OpenAI o1 System Card}, 
      author={OpenAI},
      year={2026},
      eprint={2412.16720},
      archivePrefix={arXiv},
      primaryClass={cs.AI},
      url={https://arxiv.org/abs/2412.16720}, 
}

@article{wang2022selfconsistency,
  title={Self-Consistency Improves Chain of Thought Reasoning in Language Models},
  author={Wang, Xuezhi and Wei, Jason and Schuurmans, Dale and Le, Quoc and Chi, Ed and Narang, Sharan and Chowdhery, Aakanksha and Zhou, Denny},
  journal={arXiv preprint arXiv:2203.11171},
  year={2022},
  url={https://arxiv.org/abs/2203.11171}
}

@article{chen2021codex,
  title={Evaluating Large Language Models Trained on Code},
  author={Chen, Mark and Tworek, Jerry and Jun, Heewoo and Yuan, Qiming and Pinto, Henrique Ponde de Oliveira and Kaplan, Jared and Edwards, Harri and Burda, Yuri and Joseph, Nicholas and Brockman, Greg and others},
  journal={arXiv preprint arXiv:2107.03374},
  year={2021},
  url={https://arxiv.org/abs/2107.03374}
}

@article{hendrycks2021math,
  title={Measuring Mathematical Problem Solving With the {MATH} Dataset},
  author={Hendrycks, Dan and Burns, Collin and Kadavath, Saurav and Arora, Akul and Basart, Steven and Tang, Eric and Song, Dawn and Steinhardt, Jacob},
  journal={arXiv preprint arXiv:2103.03874},
  year={2021},
  url={https://arxiv.org/abs/2103.03874}
}

@inproceedings{liu2023evalplus,
  title={Is Your Code Generated by {ChatGPT} Really Correct? Rigorous Evaluation of Large Language Models for Code Generation},
  author={Liu, Jiawei and Xia, Chunqiu Steven and Wang, Yuyao and Zhang, Lingming},
  booktitle={Thirty-seventh Conference on Neural Information Processing Systems},
  year={2023},
  url={https://openreview.net/forum?id=1qvx610Cu7}
}

@article{jain2024livecodebench,
  title={{LiveCodeBench}: Holistic and Contamination Free Evaluation of Large Language Models for Code},
  author={Jain, Naman and Han, King and Gu, Alex and Li, Wen-Ding and Yan, Fanjia and Zhang, Tianjun and Wang, Sida and Solar-Lezama, Armando and Sen, Koushik and Stoica, Ion},
  journal={arXiv preprint arXiv:2403.07974},
  year={2024},
  url={https://arxiv.org/abs/2403.07974}
}

@article{guo2025deepseekr1,
  title={{DeepSeek-R1}: Incentivizing Reasoning Capability in {LLMs} via Reinforcement Learning},
  author={{DeepSeek-AI} and Guo, Daya and Yang, Dejian and Zhang, Haowei and others},
  journal={arXiv preprint arXiv:2501.12948},
  year={2025},
  url={https://arxiv.org/abs/2501.12948}
}

@inproceedings{zhu2024deductive,
  title={Deductive Beam Search: Decoding Deducible Rationale for Chain-of-Thought Reasoning},
  author={Zhu, Tinghui and Zhang, Kai and Xie, Jian and Su, Yu},
  booktitle={First Conference on Language Modeling},
  year={2024},
  url={https://openreview.net/forum?id=S1XnUsqwr7}
}

@article{yao2023tree,
  title={Tree of Thoughts: Deliberate Problem Solving with Large Language Models},
  author={Yao, Shunyu and Yu, Dian and Zhao, Jeffrey and Shafran, Izhak and Griffiths, Thomas L. and Cao, Yuan and Narasimhan, Karthik},
  journal={Advances in Neural Information Processing Systems},
  volume={36},
  year={2023},
  url={https://arxiv.org/abs/2305.10601}
}

@inproceedings{hao2023reasoning,
  title={Reasoning with Language Model is Planning with World Model},
  author={Hao, Shibo and Gu, Yi and Ma, Haodi and Hong, Joshua Jiahua and Wang, Zhen and Wang, Daisy Zhe and Hu, Zhiting},
  booktitle={Proceedings of the 2023 Conference on Empirical Methods in Natural Language Processing},
  year={2023},
  url={https://arxiv.org/abs/2305.14992}
}

@article{azizi2026powersmc,
  title={{Power-SMC}: Low-Latency Sequence-Level Power Sampling for Training-Free {LLM} Reasoning},
  author={Azizi, Seyedarmin and Baghaei Potraghloo, Erfan and Ahmadi, Minoo and Kundu, Souvik and Pedram, Massoud},
  journal={arXiv preprint arXiv:2602.10273},
  year={2026},
  url={https://arxiv.org/abs/2602.10273}
}

@article{yang2025qwen3,
  title={{Qwen3} Technical Report},
  author={Yang, An and Li, Anfeng and Yang, Baosong and others},
  journal={arXiv preprint arXiv:2505.09388},
  year={2025},
  url={https://arxiv.org/abs/2505.09388}
}

@misc{bakouch2025smollm3,
  title={{SmolLM3}: smol, multilingual, long-context reasoner},
  author={Bakouch, Elie and Patiño, Carlos Miguel and Lozhkov, Anton and others},
  year={2025},
  howpublished={\url{https://huggingface.co/blog/smollm3}}
}

@article{ji2026scalable,
  title={Scalable Power Sampling: Unlocking Efficient, Training-Free Reasoning for {LLMs} via Distribution Sharpening},
  author={Ji, Xiaotong and Tutunov, Rasul and Zimmer, Matthieu and Bou Ammar, Haitham},
  journal={arXiv preprint arXiv:2601.21590},
  year={2026},
  url={https://arxiv.org/abs/2601.21590}
}

@inproceedings{kojima2022large,
 author = {Kojima, Takeshi and Gu, Shixiang (Shane) and Reid, Machel and Matsuo, Yutaka and Iwasawa, Yusuke},
 booktitle = {Advances in Neural Information Processing Systems},
 editor = {S. Koyejo and S. Mohamed and A. Agarwal and D. Belgrave and K. Cho and A. Oh},
 pages = {22199--22213},
 publisher = {Curran Associates, Inc.},
 title = {Large Language Models are Zero-Shot Reasoners},
 url = {https://proceedings.neurips.cc/paper_files/paper/2022/file/8bb0d291acd4acf06ef112099c16f326-Paper-Conference.pdf},
 volume = {35},
 year = {2022}
}

@inproceedings{
zelikman2022star,
title={{ST}aR: Bootstrapping Reasoning With Reasoning},
author={Eric Zelikman and Yuhuai Wu and Jesse Mu and Noah Goodman},
booktitle={Advances in Neural Information Processing Systems},
editor={Alice H. Oh and Alekh Agarwal and Danielle Belgrave and Kyunghyun Cho},
year={2022},
url={https://openreview.net/forum?id=_3ELRdg2sgI}
}

@inproceedings{
chen2024universal,
title={Universal Self-Consistency for Large Language Models},
author={Xinyun Chen and Renat Aksitov and Uri Alon and Jie Ren and Kefan Xiao and Pengcheng Yin and Sushant Prakash and Charles Sutton and Xuezhi Wang and Denny Zhou},
booktitle={ICML 2024 Workshop on In-Context Learning},
year={2024},
url={https://openreview.net/forum?id=LjsjHF7nAN}
}

@inproceedings{
cheng2025integrative,
title={Integrative Decoding: Improving Factuality via Implicit Self-consistency},
author={Yi Cheng and Xiao Liang and Yeyun Gong and Wen Xiao and Song Wang and Yuji Zhang and Wenjun Hou and Kaishuai Xu and Wenge Liu and Wenjie Li and Jian Jiao and Qi Chen and Peng CHENG and Wayne Xiong},
booktitle={The Thirteenth International Conference on Learning Representations},
year={2025},
url={https://openreview.net/forum?id=gGWYecsK1U}
}

@article{chen2021evaluating,
  title={Evaluating large language models trained on code},
  author={Chen, Mark and Tworek, Jerry and Jun, Heewoo and Yuan, Qiming and Pinto, Henrique Ponde De Oliveira and Kaplan, Jared and Edwards, Harri and Burda, Yuri and Joseph, Nicholas and Brockman, Greg and others},
  journal={arXiv preprint arXiv:2107.03374},
  year={2021}
}

@inproceedings{
lightman2024lets,
title={Let's Verify Step by Step},
author={Hunter Lightman and Vineet Kosaraju and Yuri Burda and Harrison Edwards and Bowen Baker and Teddy Lee and Jan Leike and John Schulman and Ilya Sutskever and Karl Cobbe},
booktitle={The Twelfth International Conference on Learning Representations},
year={2024},
url={https://openreview.net/forum?id=v8L0pN6EOi}
}

@article{snell2024scaling,
  title={Scaling {LLM} Test-Time Compute Optimally can be More Effective than Scaling Model Parameters},
  author={Snell, Charlie and Lee, Jaehoon and Xu, Kelvin and Kumar, Aviral},
  journal={arXiv preprint arXiv:2408.03314},
  year={2024},
  url={https://arxiv.org/abs/2408.03314}
}

@article{kimi2025k15,
  title={{Kimi k1.5}: Scaling Reinforcement Learning with {LLMs}},
  author={{Kimi Team} and Du, Angang and Gao, Bofei and others},
  journal={arXiv preprint arXiv:2501.12599},
  year={2025},
  url={https://arxiv.org/abs/2501.12599}
}

@misc{mathverify2025,
  title = {{Math-Verify}: Math Verification Library},
  author = {Kydl{\'i}{\v c}ek, Hynek},
  year = {2025},
  howpublished = {\url{https://github.com/huggingface/Math-Verify}},
  note = {Software}
}

\newpage
\appendix

\section{Broader impact}
\label{sec:impact}

Marginal sharpening improves the inference-time reasoning ability of existing
language models without additional training. This can benefit applications
where stronger reasoning is valuable, including mathematical problem solving,
code generation, proof assistance, and scientific or technical workflows. By
making test-time compute more effective, the method may also reduce the cost of
obtaining stronger answers from a fixed model.

Stronger reasoning can benefit applications such as mathematics, coding, proof
assistance, and technical workflows, but it can also amplify dual-use risks such
as malicious automation or planning. Marginal sharpening does not introduce
risks independent of the underlying model, but it can expose stronger
capabilities at inference time. Its use should therefore be paired with the same
safeguards as other reasoning-enhancement methods.

\section{Results summary}\label{app:results}

\begin{table*}[h]
\centering
\footnotesize
\setlength{\tabcolsep}{2.8pt}
\renewcommand{\arraystretch}{1.0}
\caption{Overall accuracy across datasets.}
\label{tab:overall_results}
\begin{tabular}{@{}c c l c c c c c@{}}
\toprule
Model & \(L\) & Method & \(K\) & \(S\) & MATH-500 & HumanEval+ & LiveCodeBench \\
\midrule

\multirow[c]{21}{*}{\textbf{Qwen3-4B}}
& \multirow[c]{7}{*}{2048}
& Temp. Sampling & -- & -- & 0.292 & 0.207 & 0.081 \\
& & Power Sampling & -- & -- & 0.292 & 0.482 & \textbf{0.137} \\
& & Majority Vote & -- & -- & \textbf{0.542} & 0.390 & 0.103 \\
& & Marginal & 32 & 1 & 0.522 & \textbf{0.518} & 0.131 \\
& & Marginal & 16 & 2 & 0.524 & \textbf{0.518} & 0.131 \\
& & Marginal & 8  & 4 & 0.522 & \textbf{0.518} & 0.131 \\
& & Marginal & 4  & 8 & 0.522 & 0.512 & 0.131 \\
\cmidrule(lr){2-8}

& \multirow[c]{7}{*}{4096}
& Temp. Sampling & -- & -- & 0.451 & 0.475 & 0.137 \\
& & Power Sampling & -- & -- & 0.443 & 0.756 & \textbf{0.217} \\
& & Majority Vote & -- & -- & \textbf{0.708} & 0.707 & 0.154 \\
& & Marginal & 32 & 1 & 0.700 & \textbf{0.848} & 0.200 \\
& & Marginal & 16 & 2 & 0.702 & 0.823 & 0.206 \\
& & Marginal & 8  & 4 & 0.700 & 0.829 & 0.194 \\
& & Marginal & 4  & 8 & 0.702 & 0.829 & 0.200 \\
\cmidrule(lr){2-8}

& \multirow[c]{7}{*}{8192}
& Temp. Sampling & -- & -- & 0.709 & 0.843 & 0.266 \\
& & Power Sampling & -- & -- & 0.648 & 0.890 & 0.303 \\
& & Majority Vote & -- & -- & \textbf{0.848} & 0.884 & 0.274 \\
& & Marginal & 32 & 1 & 0.840 & 0.927 & 0.303 \\
& & Marginal & 16 & 2 & 0.838 & \textbf{0.939} & 0.303 \\
& & Marginal & 8  & 4 & 0.834 & 0.902 & \textbf{0.314} \\
& & Marginal & 4  & 8 & 0.842 & 0.915 & 0.309 \\

\midrule

\multirow[c]{21}{*}{\textbf{Qwen3-8B}}
& \multirow[c]{7}{*}{2048}
& Temp. Sampling & -- & -- & 0.285 & 0.579 & 0.095 \\
& & Power Sampling & -- & -- & -- & -- & -- \\
& & Majority Vote & -- & -- & \textbf{0.602} & 0.701 & 0.103 \\
& & Marginal & 32 & 1 & 0.528 & 0.701 & 0.126 \\
& & Marginal & 16 & 2 & 0.524 & 0.707 & \textbf{0.131} \\
& & Marginal & 8  & 4 & 0.530 & 0.701 & \textbf{0.131} \\
& & Marginal & 4  & 8 & 0.528 & \textbf{0.713} & 0.120 \\
\cmidrule(lr){2-8}

& \multirow[c]{7}{*}{4096}
& Temp. Sampling & -- & -- & 0.623 & 0.784 & 0.177 \\
& & Power Sampling & -- & -- & -- & -- & -- \\
& & Majority Vote & -- & -- & \textbf{0.806} & 0.817 & 0.200 \\
& & Marginal & 32 & 1 & 0.770 & \textbf{0.848} & 0.229 \\
& & Marginal & 16 & 2 & 0.770 & 0.841 & 0.223 \\
& & Marginal & 8  & 4 & 0.770 & \textbf{0.848} & 0.223 \\
& & Marginal & 4  & 8 & 0.768 & \textbf{0.848} & \textbf{0.234} \\
\cmidrule(lr){2-8}

& \multirow[c]{7}{*}{8192}
& Temp. Sampling & -- & -- & 0.845 & 0.840 & 0.296 \\
& & Power Sampling & -- & -- & -- & -- & -- \\
& & Majority Vote & -- & -- & \textbf{0.932} & 0.854 & 0.309 \\
& & Marginal & 32 & 1 & 0.924 & 0.878 & 0.343 \\
& & Marginal & 16 & 2 & 0.928 & \textbf{0.884} & 0.343 \\
& & Marginal & 8  & 4 & 0.924 & \textbf{0.884} & 0.343 \\
& & Marginal & 4  & 8 & 0.928 & 0.860 & \textbf{0.349} \\

\bottomrule
\end{tabular}
\end{table*}

\newpage
\section{Sequence-level correction for answer decoding}
\label{app:sis}

The main decoder samples answer tokens from the locally normalized rule in
Equation~\eqref{eq:poe-token-advanced} when $S = 1$.  This is efficient, but it approximates a
sequence-level conditional.  For a fixed trace group, the exact conditional is
\begin{equation}
    \widetilde p_K(a\mid x,z_{1:K})
    =
    \frac{\prod_{i=1}^{K}\pi(a\mid x,z_i)}
    {\sum_{a'\in\A}\prod_{i=1}^{K}\pi(a'\mid x,z_i)}.
    \label{eq:app-exact-conditional}
\end{equation}
Sequential importance sampling gives a more expensive correction to this local
approximation.  At prefix \(a_{<t}\), the token-level decoder normalizes the
sequence-level product over the vocabulary,
\begin{equation}
    q_t(v\mid x,z_{1:K},a_{<t})
    =
    \frac{\prod_{i=1}^{K}\pi(v\mid x,z_i,a_{<t})}
    {\rho_t(x,z_{1:K},a_{<t})},
    \qquad
    \rho_t=\sum_{u\in\V}\prod_{i=1}^{K}\pi(u\mid x,z_i,a_{<t}).
\end{equation}
SIS restores the local normalizers omitted by this proposal by accumulating
\(\log w=\sum_t\log\rho_t\).  In logit form,
\begin{equation}
    \log \rho_t
    =
    \log\sum_{v\in\V}\exp\left(\sum_{i=1}^{K}\ell_{i,t}(v)\right)
    -
    \sum_{i=1}^{K}\log\sum_{v\in\V}\exp(\ell_{i,t}(v)).
    \label{eq:app-sis-rho-logits}
\end{equation}
The SIS variant keeps \(P\) answer particles, updates their weights after each
token, resamples when \(\ESS<P/2\), and selects a final particle according to the
normalized weights.  The case \(P=1\) follows the same sampled path as the default
token-level decoder, so the correction only changes the output when multiple
particles are used.

We use this variant as a diagnostic for the token-level approximation.  On
LiveCodeBench v6 with Qwen3-8B, \(L=8192\), \(K=4\), and \(P=8\), the sampler
performed 2.74 resampling events per problem on average, indicating that the particle
population often separated during generation.  Nevertheless, the extracted programs
were not meaningfully different from the \(P=1\) outputs.  We did not observe cases
where only one setting produced extractable code.  Among the 62 problems where both
settings produced extractable Python code, 41 outputs were identical, 15 differed only by
variable renaming or import changes that did not affect the
program structure, and the remaining 6
were manually reviewed as implementing the same logic.  The observed differences
were limited to comments, variable names, and equivalent rewrites of arithmetic,
logical, or control-flow expressions.  Table~\ref{tab:sis-diff-examples} gives
representative examples.

In this experiment, the heavier sequence-level correction mostly changed surface
form rather than task-level behavior.  This supports using the cheaper token-level
decoder as the main method.

\begin{table}[h!]
\centering
\small
\caption{Representative changed lines from \(P=1\) and \(P=8\) outputs.  The
\texttt{-} lines come from \(P=1\), and the \texttt{+} lines come from \(P=8\).}
\label{tab:sis-diff-examples}
\setlength{\tabcolsep}{4pt}
\renewcommand{\arraystretch}{1.15}
\begin{tabularx}{\textwidth}{p{0.18\textwidth}X}
\toprule
Type & Changed lines \\
\midrule
Comment &
{\scriptsize
\begin{tabular}[t]{@{}l@{}}
\texttt{-\# Initialize the count of X}\\
\texttt{+\# Count how many times X appears}
\end{tabular}} \\
\midrule
Rename &
{\scriptsize
\begin{tabular}[t]{@{}l@{}}
\texttt{- indices = positions[key]}\\
\texttt{+ lst = positions[key]}
\end{tabular}} \\
\midrule
Restructure &
{\scriptsize
\begin{tabular}[t]{@{}l@{}}
\texttt{- return len(nums) - 1 if total\_sum \% 2 == 0 else 0}\\
\texttt{+ if total\_sum \% 2 == 0:}\\
\texttt{+ \ \ \ \ return len(nums) - 1}\\
\texttt{+ else:}\\
\texttt{+ \ \ \ \ return 0}
\end{tabular}} \\
\midrule
Restructure &
{\scriptsize
\begin{tabular}[t]{@{}l@{}}
\texttt{- for i, c in enumerate(s):}\\
\texttt{- \ \ \ \ reversed\_pos = 26 - (ord(c) - ord('a'))}\\
\texttt{+ for i in range(len(s)):}\\
\texttt{+ \ \ \ \ reversed\_pos = 26 - (ord(s[i]) - ord('a'))}
\end{tabular}} \\
\bottomrule
\end{tabularx}
\end{table}


\end{document}